\documentclass[lettersize,journal]{IEEEtran}
\usepackage{amsmath,amsfonts}
\usepackage{array}
\usepackage[caption=false,font=normalsize,labelfont=sf,textfont=sf]{subfig}
\usepackage{textcomp}
\usepackage{stfloats}
\usepackage{url}
\usepackage{verbatim}
\usepackage{graphicx}
\usepackage{cite}

\usepackage{makecell,multirow}
\usepackage{threeparttable}
\usepackage{graphicx} 
\usepackage{amsmath}
\usepackage{amssymb}
\usepackage{latexsym}
\usepackage{CJK}
\usepackage{booktabs}
\usepackage{tabularx}
\usepackage{array}

\usepackage{bigstrut}
\usepackage{bm}
\usepackage{graphicx} 
\usepackage{epstopdf}

\usepackage[lined,boxed,commentsnumbered,ruled]{algorithm2e}
\newsavebox\CBox

\usepackage[pagebackref=true,breaklinks=true,colorlinks,bookmarks=false]{hyperref}

\begin{document}

\title{SVDM: Single-View Diffusion Model for Pseudo-Stereo 3D Object Detection}

\author{Yuguang Shi,
\thanks{The authors are with the School of Automation, Southeast University, Nanjing 210096, China, and also with the Key Laboratory of Measurement and Control of Complex Systems of Engineering, Ministry of Education, Nanjing 210096, China (e-mail: syg@seu.edu.cn; xblu2013@126.com).}
}

\markboth{Journal of \LaTeX\ Class Files,~Vol.~14, No.~8, August~2021}%
{Shell \MakeLowercase{\textit{et al.}}: A Sample Article Using IEEEtran.cls for IEEE Journals}

\IEEEpubid{\begin{minipage}{\textwidth}\ \\[30pt] \centering
		Copyright \copyright 20xx IEEE. Personal use of this material is permitted. 
		However, permission to use this material for any other purposes must \\ be obtained 
		from the IEEE by sending an email to pubs-permissions@ieee.org.
\end{minipage}}

\maketitle

\begin{abstract}
One of the key problems in 3D object detection is to reduce the accuracy gap between methods based on LiDAR sensors and those based on monocular cameras. A recently proposed framework for monocular 3D detection based on Pseudo-Stereo has received considerable attention in the community. However, so far these two problems are discovered in existing practices, including (1) monocular depth estimation and Pseudo-Stereo detector must be trained separately, (2) Difficult to be compatible with different stereo detectors and (3) the overall calculation is large, which affects the reasoning speed. In this work, we propose an end-to-end, efficient pseudo-stereo 3D detection framework by introducing a Single-View Diffusion Model (SVDM) that uses a few iterations to gradually deliver right informative pixels to the left image. SVDM allows the entire pseudo-stereo 3D detection pipeline to be trained end-to-end and can benefit from the training of stereo detectors. Afterwards, we further explore the application of SVDM in depth-free stereo 3D detection, and the final framework is compatible with most stereo detectors. Among multiple benchmarks on the KITTI dataset, we achieve new state-of-the-art performance. 
\end{abstract}


\section{Introduction}
\IEEEPARstart{R}{ecent} exciting solutions that generate Pseudo-Sensor representations from Monocular camera utilize pretrained monocular depth estimation network. For example, Pseudo-Stereo present an approach to infer a virtual view of a scene from a single input image, followed by applying LIGA-Stereo \cite{guo2021liga}, which is an existing Stereo-based detector. Pseudo-Stereo achieves 17.74 $AP_{3D}$ at the moderate case on the KITTI benchmark\cite{geiger2012we}.

While pseudo-stereo is conceptually intuitive, the method for generating virtual views from depth maps suffers from some limitations:
1) Although virtual views do not require real actual views in the dataset for training but still require depth ground truth to train the monocular depth estimation network, collecting large and diverse training datasets with accurate ground truth depth for supervised learning \cite{godard2019digging} is a tedious and difficult challenge in itself, so this approach inevitably increases the burden on the model.

2)The pseudo-stereoscopic approach synthesizes a pair of stereo images by forward warping. As shown in Figure 2, due to the nature of forward warping, the pseudo-right image will contain pixel artifacts that are lost due to occluded regions and in some places collisions will occur when multiple pixels will land in the same location, creating visually unpleasant holes, distortions and artifacts, thus not exploiting the potential of image-level generation for pseudo-stereoscopic 3D detection very well.

3)Stereo 3D detectors detect a variety of principles. While some of the current higher accuracy methods in the KITTI dataset ranking include a rigid accuracy depth estimation network, some geometry-based methods still have the advantages of their simplicity of principle, fast inference and scalability in low-cost scenarios. However, Feature-level Generation in pseudo-stereology is difficult to be applied directly to these methods, and has the disadvantage of limited fitness.

\begin{figure}[t]
	\centering
	\includegraphics[width=3in]{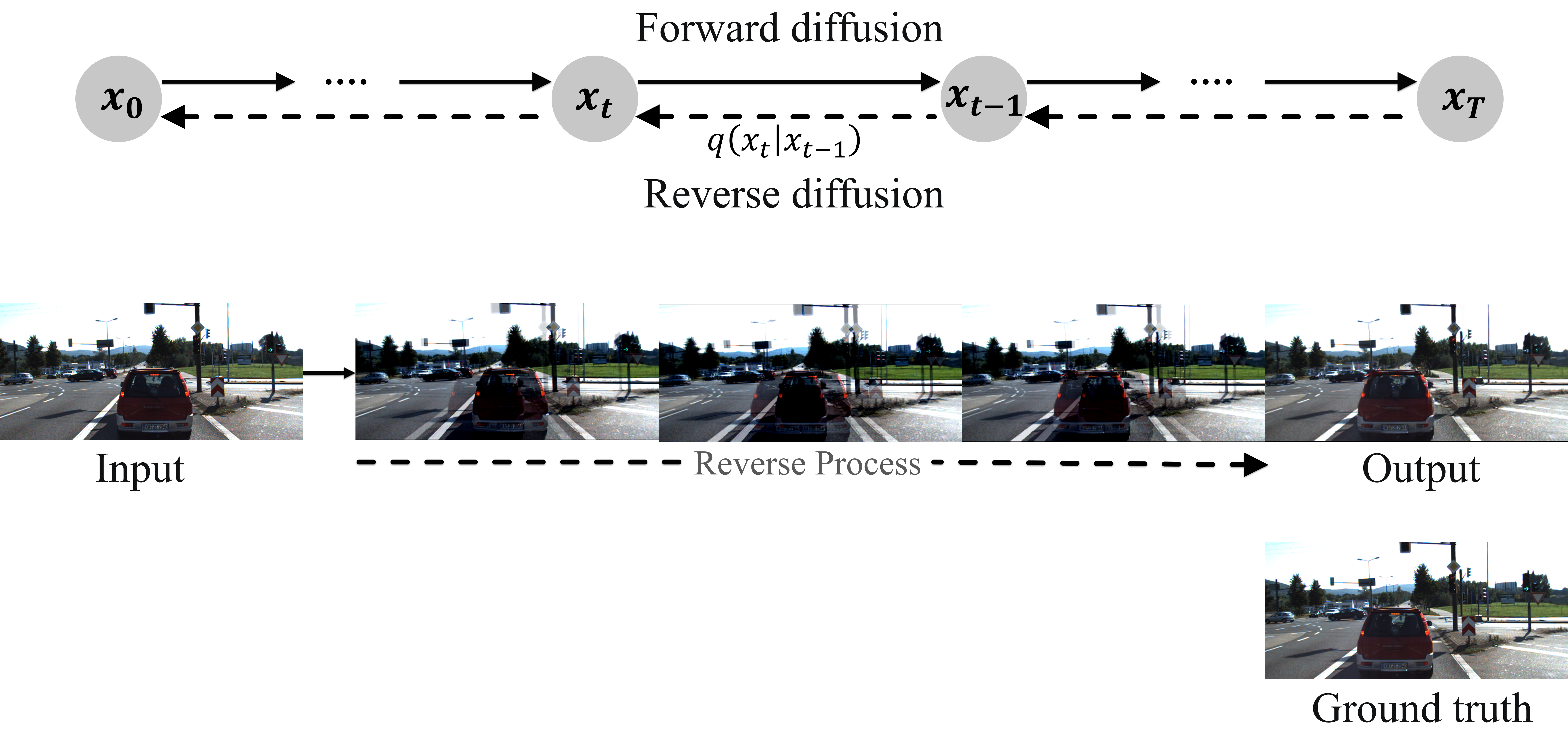}
	\caption{Overview of our SVDM framework with novel virtual view generation methods.}
	\label{fig_1}
\end{figure}

A natural question to ask, however, is whether it is possible to design a new perspective generator without depth estimation networks at the image-level? In the recent literature, diffusion not only provides significantly simpler architectures, but also offers fewer hyperparameters and simpler training steps than the notoriously difficult to train GAN. While diffusion models can generate high quality images, no study has yet demonstrated that diffusion models remain effective for the task of pseudo-view generation for stereo 3D detection.

Considering the above challenges, this study develop a new Single-View Diffusion Model (SVDM) for the high quality, spatially consistent virtual view synthesis in real scenarios.
Specifically, our method assumes that the left image in stereo views is known, replaces the Gaussian noise of the diffusion model with left image pixels during training or testing, and gradually diffuses the pixels of the right image to the full image.
Benefiting from the subtle disparity of pixels in stereo images, a few iterations can produce promising results. Note that the ground truth actual views in the dataset are only used in training.
Compared to prior work, SVDM discards the monocular depth estimation network and provides a simple end-to-end approach, so the resulting framework is compatible with most existing stereo detectors and depth estimators.  To the best of our knowledge, SVDM is the first diffusion model approach to generate virtual views from a single image input without depth estimation networks and geometric priors.

Our contributions are summarized as follows:
\begin{itemize}
	\item[$\bullet$]We introduce SVDM, an image-to-image diffusion model for pseudo-stereoscopic view generation tasks without geometric priors and depth estimation networks. SVDM provides competitive results compared to current monocular 3D detectors on the KITTI-3D benchmark.
	\item[$\bullet$]We introduce three new diffusion model approaches for transforming new view generation tasks into image-to-image translation tasks.
	\item[$\bullet$]We introduce ConvNeXt-UNet, a new UNet architectural variant for new view synthesis, showing that architectural changes are crucial for high-fidelity results.
\end{itemize}

\begin{figure}[t]
	\centering
	\includegraphics[width=8cm]{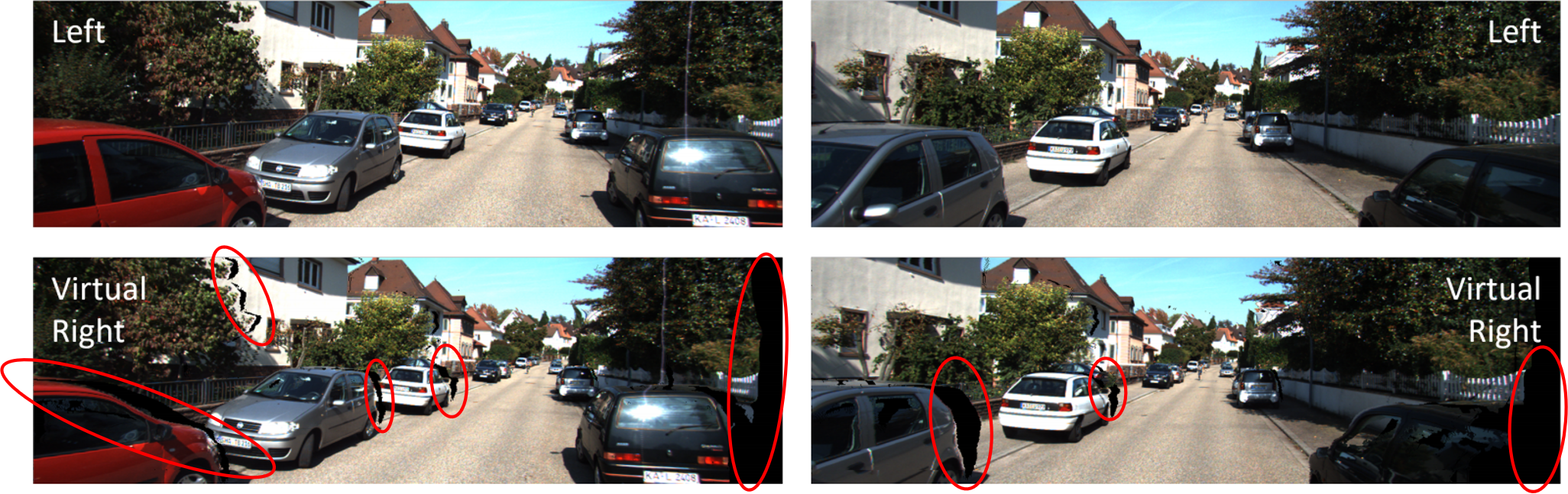}
	\caption{Left image (top) and the generated virtual right image (bottom) using our image-level virtual view generation method.}
	\label{fig_2}
\end{figure}

\section{RELATED WORK}

In this part, we briefly review the literature on monocular 3D object detection ,view synthesis and  diffusion models in recent years.

\noindent{\bf Monocular 3D detection:} According to the input representation, monocular 3D detectors are roughly divided into image-based methods and depth-based methods. Image-based methods focus on reducing the dimensionality of 3D problems to 2D or 2.5D problems to save the amount of calculation from depth estimation networks. A few works \cite{mousavian20173d,zhang2021learning,simonelli2020towards,lu2021geometry} introduce perspective projection model to calculate depth information, but projection process introduces the error amplification problem, hurting the performance of deep inferences. M3D-RPN \cite{brazil2019m3d} is the first anchor-based method, these 2D and 3D anchor boxes are placed on the image pixels, the depth parameter is encoded by projecting the 3D center location, and some works \cite{brazil2020kinematic,kumar2021groomed,liu2021ground} have tried to improve this method. 
CenterNet \cite{centernet} is an anchor-free 2D detector that has a profound impact on 3D detection by applying multiple heads to predict 3D properties, and a series of improved methods \cite{RTM3D,SMOKE,ma2021delving,li2021monocular,chen2020monopair,zhang2021objects,yang2021lite,li2022diversity} based on point features have been proposed.

Inspired by the success of monocular depth estimation networks, performances of state-of-the-art depth-based methods aggregate image and depth features to obtain depth-aware features due to the geometric information loss during imagery projection. Mono3D \cite{chen2016monocular} exploits segmentation, context and location priors to generate 3D proposals. MonoGRNet \cite{qin2019monogrnet} employs sparse supervision to directly predict object center depth, and optimizes 3D information through multi-task learning. D4LCN \cite{ding2020learning} proposes depth-guided dynamic expansion local convolutional network, which address the problem of the scale-sensitive and meaningless local structure in existing works. DDMP \cite{wang2021depth} alleviates the challenge of inaccurate depth priors by combining multi-scale depth information with image context. A line of Transformer-based methods \cite{zhang2022monodetr,huang2022monodtr} have a similar pipeline in that encode depth information into a 2D detector named detr. 

Another family of Pseudo-LiDAR architecture such as \cite{wang2019pseudo,ma2020rethinking,reading2021categorical,simonelli2021we,ma2019accurate,bao2019monofenet,park2021pseudo}, back-projects depth map pixels into point-cloud 3D coordinates, and then apply ideas of point-cloud based detector. These methods narrow the accuracy gap between monocular and lidar and can be continuously improved by subsequent depth estimation networks and point-cloud based detectors. RefinedMPL \cite{vianney2019refinedmpl} uses PointRCNN \cite{shi2019pointrcnn} for point-wise feature learning in a supervised or an unsupervised scheme from pseudo point clouds prior. AM3D \cite{ma2019accurate} uses a PointNet \cite{qi2017pointnet} backbone for point-wise feature extraction from pseudo point clouds, and employs a multi-modal fusion block to enhance the point-wise feature learning. MonoFENet \cite{bao2019monofenet} enhances the 3D features from the estimated disparity for monocular 3D detection. Decoupled-3D \cite{cai2020monocular} recovers the missing depth of the object using the coarse depth from 3D object height prior with the BEV features that are converted from the estimated depth map. Pseudo-Stereo \cite{chen2022pseudo} further proposes the intermediate stereo representation for converting monocular imagery data to Pseudo-LiDAR signal. Despite the improvement of Pseudo-Stereo, its novel virtual view synthesis methods have certain limitations in the scope of application of stereo detectors.

\begin{figure*}[t]
	\centering
	\includegraphics[width=18cm,height=8.346cm]{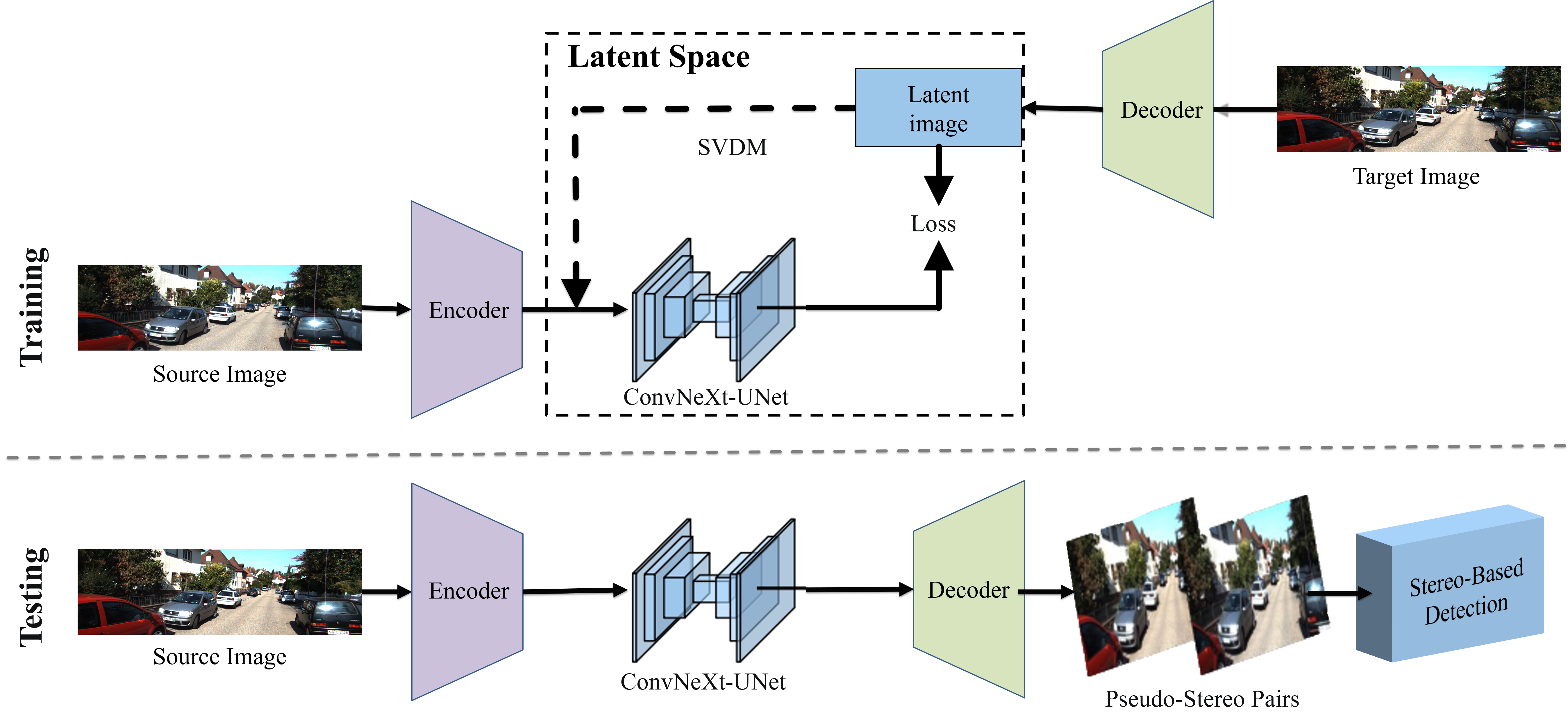}
	\caption{Overview of our virtual view generation methods.}
	\label{fig_3}
\end{figure*}

\noindent{\bf Novel View Synthesis:} 
Novel view synthesis is a highly ill-posed problem that focuses on generating new views of scenes. The classic work uses the depth map to forward warp the image pixels into the novel views. In order to overcome the challenging problem that large quantities of ground-truth depth data are difficult to obtain, some self-supervised methods \cite{xie2016deep3d,garg2016unsupervised,godard2017unsupervised,mahjourian2018unsupervised,godard2019digging,shu2020feature,peng2021excavating,he2022ra} only use stereo raw images to train a model. To deal with holes, cracks, and blurs, there are also attempts to study the improvement of the quality of the synthetic images. Tulsiani et al. \cite{tulsiani2018layer} propose a layed depth image (LDI) 3D representation to capture the texture and depth of the foreground and background. Stereo Magnification: Learning view synthesis using multiplane images\cite{zhou2018stereo}
Chen et al. propose a learning framework based on multiplane images (MPIs), and a series of MPI-based methods \cite{brazil2020kinematic,kumar2021groomed,liu2021ground} have been developed. 
Inspired by NeRF \cite{mildenhall2021nerf} with MPI \cite{tucker2020single}, MINE \cite{li2021mine} achieve competitive novel view images and depth maps from a single input image. To reduce the influence of parallax on the network, SivsFormer \cite{zhang2022sivsformer} designed a warping and occlusion handing module to improve the quality of the synthetic images. Nonetheless, these methods heavily rely on specially designed pipelines or explicit geometric models. Recently, denoising diffusion models demonstrating great potential in various computer vision fields including super-resolution \cite{saharia2022image,sahak2023denoising}, image generation \cite{dhariwal2021diffusion,gu2022vector,ho2022cascaded,meng2021sdedit,nichol2021improved,peebles2022scalable,wang2022sindiffusion,liu2022flow}, object detection and segmentation \cite{amit2021segdiff,baranchuk2021label,brempong2022denoising,chen2022diffusiondet}, etc.

In this work, we consider the particular image generation task of P-stereo 3D detection, taking full advantage of binocular stereo lenses and exploiting diffusion models, and propose a novel geometrically free viewpoint generation framework, which we call SVDM. Our framework can be applied to both offline and online generation based on different diffusion model methods, and achieves good generation results without depth images and explicit geometric priors.

\section{THE PROPOSED METHOD}

\subsection{Preliminaries}

\subsubsection{Stereo 3D Detector}
Stereo 3D object detection is a unique branch of 3D detection that aims to predict the location, size, orientation and category of an object in 3D space using only a stereo camera sensor. According to the type of training data, stereo-imagery-based methods can be generally divided into three types. The first type solely requires stereo images with corresponding annotated 3D bounding boxes. According to the type of training data, stereo image-based methods can be generally classified into three types. The first type only requires stereo images with corresponding annotated 3D bounding boxes, and this approach wants to take full advantage of the geometric relationships and pixel constraints of stereo images without using depth estimation networks, represented by TLNet \cite{qin2019triangulation}, Stereo R-CNN \cite{Stereor-cnn} and Stereo CenterNet \cite{Stereor-cnn}. The second type requires an additional depth map to train the data, and representative methods are pseudo-LiDAR family\cite{wang2019pseudo,you2019pseudo,qian2020end}, IDA-3D \cite{IDA}, YOLOStereo3D, etc. The third type is called Volume-based method, which recodes 3D objects and locates 3D objects from 3D feature volume, represented by DSGN series methods and LIGA-Stereo. For a fair comparison and to demonstrate the scalability of our approach, we used three methods, stereo-rcnn, LIGA-Stereo and stereoyolo, as our base stereo 3D detection system, and the generated pseudo-stereo images were fed to all three methods.

\subsubsection{Denoising Diffusion Probabilistic Models}
A T-step Denoising Diffusion Probabilistic Model (DDPM) \cite{ho2020denoising} consists of two processes: the forward process (also referred to as diffusion process), and the reverse inference process. 

The forward process $q ( x _ { t } | x _ { t - 1 } )$ is adding noise to the picture. For example, give a picture $x _ { 0 }$, the forward process adds Gaussian noise to it through $T$ times of  accumulation to obtain $x _ { 1 } , x _ { 2 } , \cdots , x _ { T }$. The step sizes are controlled by a variance schedule$\{ \beta _ { t } E ( 0 , 1 ) \} _ { t = 1 } ^ { T }$. Each time $t$ in the forward process is only related to time $t - 1$, so it can be regarded as a Markov process.
\begin{eqnarray}
q\left(\mathbf{x}_{t} \mid \mathbf{x}_{t-1}\right)=\mathcal{N}\left(\mathbf{x}_{t} ; \sqrt{1-\beta_{t}} \mathbf{x}_{t-1}, \beta_{t} \mathbf{I}\right)
\end{eqnarray}
\begin{eqnarray}
q\left(\boldsymbol{x}_{1}, \ldots, \boldsymbol{x}_{T} \mid \boldsymbol{x}_{0}\right)=\prod_{t=1}^{T} q\left(\boldsymbol{x}_{t} \mid \boldsymbol{x}_{t-1}\right)
\end{eqnarray}

If the forward process is the process of adding noise, then the reverse process is 
denoising process of diffusion. If the reversed distribution: $q ( x _ { t - 1 } | x _ { t } )$ is obtained, we can restore the original distribution from the complete standard Gaussian distribution. Unfortunately, we cannot easily estimate $q ( x _ { t - 1 } | x _ { t } )$ because it needs to use the entire dataset and therefore we need to learn a model $p _ {\theta }$ to approximate these conditional probabilities in order to run the reverse diffusion process.
\begin{eqnarray}
p_{\theta}\left(\boldsymbol{x}_{0}, \ldots, \boldsymbol{x}_{T-1} \mid \boldsymbol{x}_{T}\right)=\prod_{t=1}^{T} p_{\theta}\left(\boldsymbol{x}_{t-1} \mid \boldsymbol{x}_{t}\right)
\end{eqnarray}
\begin{eqnarray}
p _ { \theta } ( x _ { t - 1 } ) ( x _ { t } ) = N ( x _ { t - 1 } ; H _ { \theta } ( x _ { t } , t ) , \sum _ { \theta } ( x _ { t } , t ) )
\end{eqnarray}

In one sentence, the diffusion model is to destroy the training data by continuously adding Gaussian noise, and then restore the data by learning the reverse denoising process. After training, the Diffusion Model can be used to pass randomly sampled noise into the model, and generate data through the learned denoising process.

The training objective of DDPM is to optimize the Evidence Lower Bound (ELBO). Finally, the objective can be simplified as to optimize:

\begin{eqnarray}
\mathbb{E}_{\boldsymbol{x}_{0}, \boldsymbol{\epsilon}}\left\|\boldsymbol{\epsilon}-\boldsymbol{\epsilon}_{\theta}\left(\boldsymbol{x}_{t}, t\right)\right\|_{2}^{2}
\end{eqnarray}

where $\epsilon$ is the Gaussian noise in ${x}_{t}$ which is equivalent to
$\Delta _ { x _ { t } } \ln q ( x _ { t } | x _ { 0 } )$, $\boldsymbol{\epsilon}_{\theta}$ is the model trained to estimate $\epsilon $. Most conditional diffusion models maintain the forward process and directly inject the condition into the training objective:

\begin{eqnarray}
\mathbb{E}_{\boldsymbol{x}_{0}, \boldsymbol{\epsilon}}\left\|\boldsymbol{\epsilon}-\boldsymbol{\epsilon}_{\theta}\left(\boldsymbol{x}_{t}, \boldsymbol{y}, t\right)\right\|_{2}^{2}
\end{eqnarray}

Since $p ( x _ { t } | y )$ dose not obviously appear in the training objective, it is difficult to guarantee the diffusion can finally reaches the desired conditional distribution Except for the conditioning mechanism, Latent Diffusion Model (LDM) takes the diffusion and inference processes in the latent space of VQGAN, which is proven to be more efficient and generalizable than operating on the original image pixels.

\subsection{Single-View Diffusion Model}
The proposed framework views the new view generation task as an image-to-image translation (I2I) task based on diffusion model, which takes a single source image captured by a camera as input. And aim to generate a predicted view. While standard diffusion models contaminate and restore images with Gaussian noise, in this work we consider three novel diffusion methods for establishing a mapping between the input and output domains. The pipeline of the proposed method is shown in Fig. 3, which Our three diffusion model methods are presented in Section 3.2, including the Gaussian noise operator in Section 3.2.a, the view image operator in Section 3.2.b, and the one-step generation in Section 3.2.c. 

\subsubsection{Gaussian Noise Operator}
For diffusion probabilistic models used for an image generation task, the forward diffusion process of the model adds noise to a clean source image until the image is standard normal distribution, and the reverse inference process maps the noise back to the image, however this approach is not suitable for the vast majority of downstream tasks. To learn the translation between two different view domains directly in the bidirectional diffusion process of the diffusion model, following BBDM \cite{li2023bbdm}, we use the Brownian Bridge diffusion process instead of the existing DDPM methods.

A Brownian bridge is a continuous-time stochastic model in which the probability distribution during the diffusion process is conditioned on the starting and ending states. Specifically, the state distribution at each time step of a Brownian bridge process starting from point $x _ { 0 }\sim q _ { d a t a } ( x _ { 0 } )$ at $t = 0$ and ending at point $x _ { T }$ at $t = T$ can be formulated as:

\begin{eqnarray}
q _ { BB } ( x _ { t } | x _ { 0 } , y ) = \mathcal{N}( x _ { t } ; ( 1 - m_ { t }) x _ { 0 } +  m_ { t }y ,\delta_ {t}I )
\end{eqnarray}

where $m _ { t } = \frac { t } { T }$, $\delta_ {t}$ is the variance, to avoid the problem that large variance may cause the framework to fail to train properly, a schedule of variance for Brownian Bridge diffusion process can be designed as:

\begin{eqnarray}
\begin{aligned}
\delta_{t} & =1-\left(\left(1-m_{t}\right)^{2}+m_{t}^{2}\right) \\
& =2\left(m_{t}-m_{t}^{2}\right)\\
& = 2s\left(m_{t}-m_{t}^{2}\right)
\end{aligned}
\end{eqnarray}

where $s$ is the scaling factor set to $1$ by default, and the value of $s$ is adjusted to control the diversity of samples. The complete forward process can be described as follows, when $t=0$, we get $m_ {0}=0$ with mean equal to $x_ {0}$ and probability 1 and variance $\delta_ {t}=0$. When the diffusion process reaches the target $t = T$, we get $m_ {T}=1$, $x_ {T}=y$ and variance $\delta_ {T}=0$. The intermediate state $x_ {t}$ is calculated in discrete form as follows:

\begin{eqnarray}
x _ { t } = ( 1 - m _ { t } ) x _ { 0 } + m _ { t } y + \sqrt { \delta} \epsilon _ { t }
\end{eqnarray}
\begin{eqnarray}
x _ { t - 1 } = ( 1 - m _ { t - 1 } ) x _ { 0 } + m _ { t + } y + \sqrt {\delta _ { t - 1 }} \epsilon _ { t-1}
\end{eqnarray}

where $\epsilon _ { t , } \epsilon _ { t - 1 } \sim N ( 0 , I )$. The expression of $x _ { 0 }$ in equation (6) is substituted into equation (7) to obtain the transition probability $q _ { BB } ( x _ { t } | x _ { t - 1 } , y )$:

\begin{eqnarray}
\begin{array}{l}
	q_{B B}\left(\boldsymbol{x}_{t} \mid \boldsymbol{x}_{t-1}, \boldsymbol{y}\right)=\mathcal{N}\left(\boldsymbol{x}_{t} ; \frac{1-m_{t}}{1-m_{t-1}} \boldsymbol{x}_{t-1}\right. \\
	\left.\quad+\left(m_{t}-\frac{1-m_{t}}{1-m_{t-1}} m_{t-1}\right) \boldsymbol{y}, \delta_{t \mid t-1} \boldsymbol{I}\right)
\end{array}
\end{eqnarray}
where $\delta_{t \mid t-1}$ is calculated by $\epsilon _ { t }$ as:

\begin{eqnarray}
\delta _ { t | t - 1 } = \delta _ { t } - \delta_ { t - 1 } \frac { ( 1 - m _ { t } ) ^ { 2 } } { ( 1 - m _ { t - 1 } ) ^ { 2 } }
\end{eqnarray}

In the reverse process of our method, the diffusion process starts from a source image sampled from a known view, and step by step to get the target view distribution. That is, predicting $x_ { t - 1 }$ based on $x_ { t}$.

\begin{eqnarray}
p_{\theta}\left(\boldsymbol{x}_{t-1} \mid \boldsymbol{x}_{t}, \boldsymbol{y}\right)=\mathcal{N}\left(\boldsymbol{x}_{t-1} ; \boldsymbol{\mu}_{\theta}\left(\boldsymbol{x}_{t}, t\right), \tilde{\delta}_{t} \boldsymbol{I}\right)
\end{eqnarray}

where $\boldsymbol{\mu}_{\theta}\left(\boldsymbol{x}_{t}, t\right)$ is the predicted mean value of the noise, which needs to be learned by a neural network with parameter $\theta$ based on the maximum likelihood criterion. $\tilde{\delta}_{t}$ is the variance of noise at each step, which does not have to be learned and is expressed in the analytic form as $\delta _ { t } = \frac { \delta _ { t | t - 1 } \cdot \delta _ { t - 1 } } { \delta _ { t } }$. The whole training process and sampling process are summarized in Algorithm 1 and Algorithm 2.

\begin{algorithm}
	\label{alg1}
	
	\caption{Training for BBDM.}
	\LinesNumbered 
	    \Repeat{converged}{
		paired data  $x _ { 0 } \sim q ( x _ { 0 } ) , y \sim q ( y )$\;
		timestep  $t \sim U n i f o r m ( 1 , \cdots , T )$\;
		Gaussian noise  $\epsilon\sim \mathcal{N} ( 0 , I )$\;
		Forward diffusion $x _ { t } = ( 1 - m _ { t } ) x _ { 0 } + m _ { t } y + \sqrt { \delta _{t} }\epsilon $\;
		Take gradient descent step on  $\bigtriangledown_ { \theta } | | m _ { t } ( y - x _ { 0 } ) + \sqrt { \delta _{t} }\epsilon  - \operatorname { \epsilon _{\theta } } ( x _ { t } , t ) | | ^ { 2 }$\;
	}

\end{algorithm}
\begin{algorithm}
	\label{alg2}
	
	\caption{Sampling for BBDM.}
	\LinesNumbered 
	sample conditional input $\boldsymbol{x}_ { T } = \boldsymbol{y} \sim q ( \boldsymbol{y})$\;
	\For{$t=T;t\ge  1;t--$}{
		\If{$t >  1$}{$z \sim N ( 0 , I )$\;}
		\Else{$z = 0$\; }		
		$\boldsymbol{x}_{t-1}=c_{x t} \boldsymbol{x}_{t}+c_{y t} \boldsymbol{y}-c_{\epsilon t} \epsilon_{\theta}\left(\boldsymbol{x}_{t}, t\right)+\sqrt{\tilde{\delta}_{t}} \boldsymbol{z}$
	}
\Return{$x_{ 0 }$}
\end{algorithm}

\subsubsection{View Image Operator}
However, the Brownian Bridge diffusion process introduces additional hyperparameters that increase the flow and complexity of the experiment. To overcome this, we propose a View Image Operator-based method, specifically, we treat the target image as a special kind of noise and iteratively convert the target image to the source image.
Given initial state x0 and destination state y, the intermediate state xt can be written in discrete form as follows:

\begin{eqnarray}
x _ { t } = \sqrt { \alpha _ { t } }x  + \sqrt { 1 - \alpha _ { t }} z 
\end{eqnarray}
Note this is essentially the same as the noising procedure, but instead of adding noise we are adding a progressively higher weighted Novel view image. In order to sample from the learned distribution, we use Algorithm 3 to reverse the View-Image transformation. Following \cite{nichol2021improved} ,this method simply uses a schedule in terms of $\alpha _ { t }$ to interpolate.

\begin{eqnarray}
\alpha _ { t } =\frac { f ( t ) } { f ( 0 ) } , f ( t ) = \cos ( \frac { t / T + s } { 1 + s } \cdot \frac { \pi } { 2 } ) ^ { 2 }
\end{eqnarray}

\begin{figure}[t]
	\begin{center}
		\includegraphics[width=8cm]{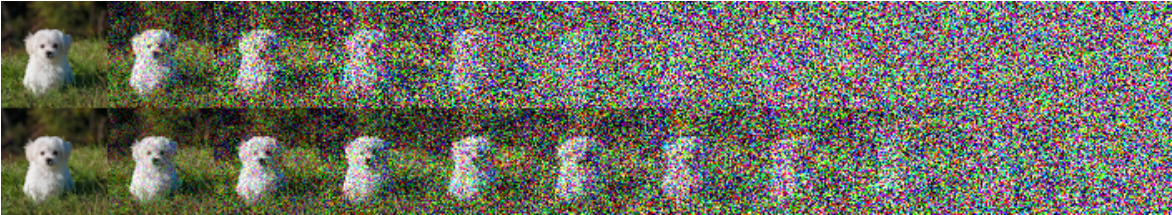}
	\end{center}
	\caption{Latent samples from linear (top) and cosine (bottom) schedules respectively at linearly spaced values of t from 0 to T.}
	\label{c}
\end{figure}

\begin{figure}[t]
	\begin{center}
		\includegraphics[width=8cm]{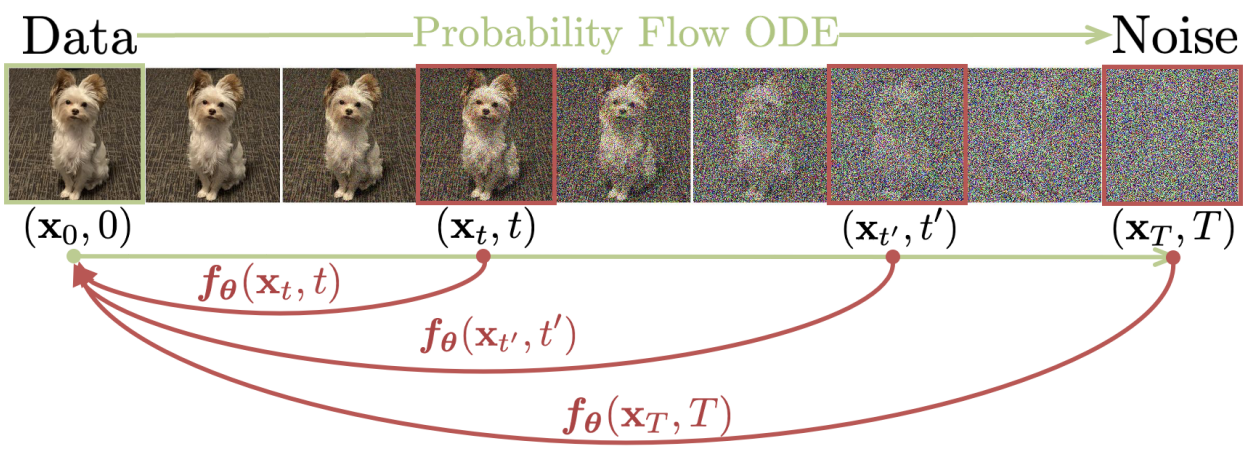}
	\end{center}
	\caption{ Given a Probability Flow (PF) ODE that smoothly converts data to noise, we learn to map any point (e.g., $X _ { t }$, $X_{t^{\prime}}$ , and $X _ { T }$ ) on the ODE trajectory to its origin (e.g., $X _ { 0}$) for generative modeling. Models of these mappings are called consistency models, as their outputs are trained to be consistent for points on the same trajectory.
 }
	\label{d}
\end{figure}

\begin{figure}[t]
	\begin{center}
		\includegraphics[width=8cm]{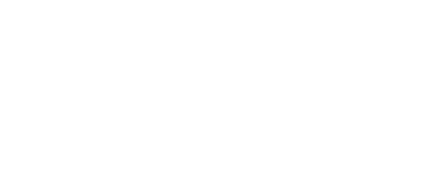}
	\end{center}
	\caption{ConvNeXt-UNet Architecture – We modify the typical UNet architecture used by recent work on diffusion models to accomodate 3D novel view synthesis. }
	\label{d}
\end{figure}

Where $s = 0.008$. The difference between linear and cosine schedules is shown in Figure 4, where it can be seen that in the later stages of linear scheduling it is almost purely a target view, while cosine scheduling adds target views more slowly. 

\begin{algorithm}
	\label{alg3}
	
	\caption{View-Image Operator Sampling.}
	\LinesNumbered 
	\KwIn{Source Image $x _ { t }$;}
	\For{$i=0;i\le l;i++$}{
		$x _ { 0 } \leq f ( x _ { s } , s )$\;
		$x_{s-1}=x_{ s } - D ( x _ { 0 } , s ) + D (x_{0},s-1)$\;}
	\Return{Target Image $x_{ 0 }$}
\end{algorithm}

\subsubsection{Accelerated Sampling And One-Step Generation}
Despite their high-quality generation performance, DPMs still suffer from their
slow sampling as they generally need hundreds or thousands of sequential function
evaluations (steps) of large neural networks to draw a sample. In recent years, several studies have been devoted to reducing the steps of DPMs, such as \cite{song2020denoising,lu2022dpm,lu2022dpm,liu2022flow,song2023consistency}, .etc. For pseudo-stereo 3D detection, the slow new view generation speed can greatly hinder the detection and deployment, so we propose two schemes in this section for accelerating the inference process of SVDM. One is a method that adds a high-order solver for the guided sampling of DPMs, and the other is to improve the one-step generation method.
\subsubsection*{\bf Accelerated Sampling}Similar to the basic idea of DDIM \cite{song2020denoising}, the inference processes of BBDM can be accelerated by utilizing a nonMarkovian process while keeping the same marginal distributions as Markovian inference processes.

Now, given a sub-sequence of $[ 1 : T ]$ of length $\mathcal{S}$ $\{ T _ { 1 } , T _ { 2 } , \cdots , T _ { S } \}$, the inference process can be defined by a subset of the latent variables $x _ { 1 : T }$ , which is $\{ x _ { T _ { 1 } } , x _ { T _ { 2 } } , \cdots , x _ { T s } \}$,
$$
\begin{array}{l}
q_{B B}\left(\boldsymbol{x}_{\tau_{s-1}} \mid \boldsymbol{x}_{\tau s}, \boldsymbol{x}_{0}, \boldsymbol{y}\right)=\mathcal{N}\left(\left(1-m_{\tau_{s-1}}\right) \boldsymbol{x}_{0}+m_{\tau_{s-1}} \boldsymbol{y}+\right. \\
\left.\sqrt{\delta_{\tau_{s-1}}-\sigma_{\tau_{s}}^{2}} \frac{1}{\sqrt{\delta_{\tau_{s}}}}\left(\boldsymbol{x}_{\tau_{s}}-\left(1-m_{\tau_{s}}\right) \boldsymbol{x}_{0}-m_{\tau_{s}} \boldsymbol{y}\right), \sigma_{\tau_{s}}^{2} \boldsymbol{I}\right)
\end{array}
$$
\subsubsection*{\bf One-Step Generation}In this section, our objective is to create generative models that facilitate efficient, single-step generation without sacrificing important advantages of iterative refinement. Following consistency models \cite{song2023consistency}, these advantages include the ability to trade-off compute for sample quality when necessary, as well as the capability to perform zeroshot data editing tasks. As illustrated in Fig. 5, we build on top of the probability flow (PF) ordinary differential equation (ODE) in continuous-time diffusion models \cite{song2020score}, whose trajectories smoothly transition the data distribution into a tractable noise distribution. We propose to learn a model that maps any point at any time step to the trajectory is starting point. A notable property of our model is self-consistency: points on the same trajectory map to the same initial point. 

Consistency models allow us to generate data samples (initial points of ODE trajectories, e.g., $\boldsymbol{x}_{0}$ in Fig. 5) by converting random noise vectors (endpoints of ODE trajectories, e.g.,  $\boldsymbol{x}_{T}$ in Fig. 5) with only one network evaluation. Importantly, by chaining the outputs of consistency models at multiple time steps, we can improve sample quality and perform zero-shot data editing at the cost of more compute, similar to what iterative refinement enables for diffusion models. eliminates the need for a
pre-trained diffusion model altogether, Consistency models allowing us to train a consistency model in isolation. This approach situates consistency models as an independent family of generative models. More formula derivation, please see the original paper.

\subsection{Model Architecture}
Following the Latent diffusion model (LDM) \cite{rombach2022high}, SVDM performs generation learning in the latent space instead of raw pixel space to reduce computational costs. In the following, we briefly recall LDM and then introduce our ConvNeXt-UNet on the latent input.

LDM employs a pretrained VAE encoder $\mathbf{E} $ to encode an image $v \in R ^ { 3 \times H \times W }$to a latent embedding $z = E ( v ) \in R ^ { c \times h \times w }$. It gradually adds noise to $z$ in the forward process and then denoises to predict $z$ in the reverse process. Finally, LDM uses a pre-trained VAE decoder $\mathbf{D} $ to decode $z$ into a high-resolution image $v = \mathbf{D}( z )$. Both VAE encoder and decoder are kept fixed during training and inference. Since $h$ and $w$ are smaller than $H$ and $W$, performing the diffusion process in the lowresolution latent space is more efficient compared to the pixel space. In this work, we adopt the efficient diffusion process of LDM. Given an image $I_{A}$ sampled from domain A, we can first extract the latent feature $L_{A}$, and then the proposed SVDM process will map $L_{A}$ to the corresponding latent representation $L _ { A \rightarrow B }$ in domain B. Finally, the translated image $I _ { A \rightarrow B }$ can be generated by the decoder of the pre-trained VQGAN \cite{esser2021taming}.

As shown in Fig. 5, the SVDM model simply connects two images along the channel dimensions and uses the standard U-Net \cite{ronneberger2015u} architecture with a ConvNeXt residual block \cite{liu2022convnet,woo2023convnext} for upsampling and downsampling the activations, reaching large receptive fields with stacked convolutions to take advantage of context information in images. This “Concat-UNet” has found significant success in prior work of image-to-image diffusion models. In addition, we introduce multiple attention blocks at various resolutions, in light of the discovery that global interaction significantly improves reconstruction quality on much larger and more diverse datasets at higher resolutions.

\subsection{Loss Functions}

There are four terms in the loss function: RGB L1 loss $\mathcal{L}_{\text {1 }}$, RGB SSIM loss $\mathcal{L}_{\text {ssim }}$, and the perceptual loss $\mathcal{L}_{\text {latent }}$ from \cite{rombach2022high}. The total loss $\mathcal{L}$ is given by:

\begin{eqnarray}
\mathcal{L}=\lambda_{\mathrm{L} 1} \mathcal{L}_{\mathrm{L} 1}+\lambda_{\text {ssim }} \mathcal{L}_{\text {ssim }}++\lambda_{1 a t e n t} \mathcal{L}_{1 a t e n t}
\end{eqnarray}
where $ \lambda _ { L 1 }, \lambda _ { s s i m }$ and $\lambda _ {1 a t e n t }$ are hyperparameters to weigh the respective loss term.

\subsubsection{RGB L1 and SSIM Loss.}
The L1 and SSIM \cite{wang2004image} losses:
\begin{eqnarray}
\mathcal{L}_{\mathrm{L} 1} = \frac{1}{3HW} \sum \left | \hat{I} _ { t g t  }-  I _ { t g t } \right | 
\end{eqnarray}
\begin{eqnarray}
\mathcal{L} _ { s s i m } = 1 - S SI M ( \hat{I} _ { t g t  }, I _ { t g t } )
\end{eqnarray}
are to encourage the synthesized target image $\hat{I} _ { t g t  }$ to match the ground truth Itgt. Both $\hat{I} _ { t g t  }$ and $I _ { t g t }$ are 3-channel RGB images of size $H \times  W$.

\subsubsection{Perceptual Loss.}Perceptual compression model is based on previous
work \cite{esser2021taming} and consists of an autoencoder trained by combination of a perceptual loss \cite{zhang2018unreasonable} and a patch-based \cite{isola2017image} adversarial objective \cite{dosovitskiy2016generating,esser2021taming,yu2021vector}. This ensures that the reconstructions are confined to the image manifold by enforcing local realism and avoids bluriness introduced by relying solely on pixel-space losses such as L2 or L1 objectives.

\begin{eqnarray}
\mathcal{L}_{\text {latent }}=\frac{1}{2} \sum_{\mathrm{j}=1}^{J}\left[\left(u_{j}^{2}+\sigma_{j}^{2}\right)-1-\log \sigma_{j}^{2}\right]
\end{eqnarray}

\begin{figure*}[t]
	\centering
	\includegraphics[width=18cm,height=10.2cm]{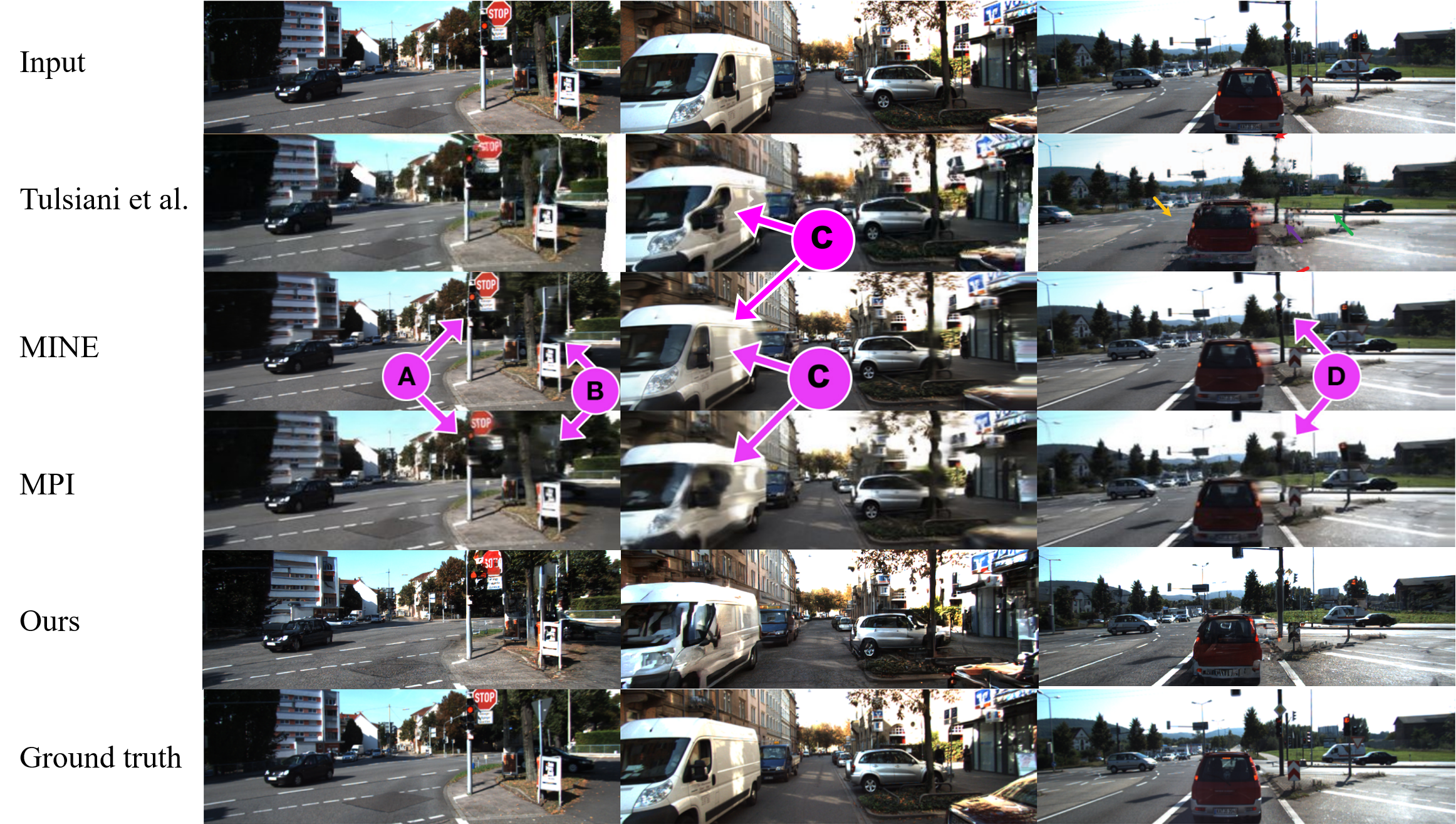}
	\caption{Qualitative comparison on KITTI.}
	\label{fig_5}
\end{figure*}
\begin{table*}[htbp]
	\centering
	\scriptsize
	\caption{View synthesis on KITTI dataset. }
	
	\begin{tabular}{c|cccc|ccc}
		\toprule
		& Train Res. & N     & Pre-trained & Depth Smoothess & LPIPS$\downarrow $ & SSIM$\uparrow$ & PSNR$\uparrow$ \\
		\midrule
		DAM-CNN \cite{xie2016deep3d} & 768x256 & NA    & NA    & Y     & 0.205 & 0.598 & 17.3 \\
		Tulsiani et. al. \cite{tulsiani2018layer} & 768x256 & NA    & NA    & NA    & -     & 0.572 & 16.5 \\
		MPI \cite{tucker2020single} & 768x256 & 32    & NA    & NA    & -     & 0.733 & 19.5 \\
		MINE \cite{li2021mine} & 768x256 & 32    & Y     & Y     & 0.112 & 0.822 & 21.4 \\
		MINE \cite{li2021mine} & 768x256 & 64    & Y     & Y     & 0.108 & 0.820 & 21.3 \\
		SVDM  & 768x256 & NA    & NA    & NA    & 0.257 & 0.768 & 21.5 \\
		\bottomrule
	\end{tabular}%
	
	\label{tg}%
\end{table*}%
\section{EXPERIMENTS}
\subsection{Datasets}
For novel view synthesis and 3D detection, we perform both quantitative and qualitative comparisons with state-of-the-art methods on the KITTI datasets \cite{geiger2012we}.

\subsubsection{View Synthesis}
According to the suggestions of Tulsiani et al. in \cite{tulsiani2018layer}, we randomly choose 22 sequences from the whole data for training, and the remaining 8 sequences are equally divided by validation set and test set. The training set contains about 6000 stereo pairs, the test sequences set contains 1079 image pairs, and the images contain a large number of occlusions, such as cars, pedestrians, traffic lights, etc.We use the left camera image as the source image and the other as the target view image. Following \cite{tucker2020single}, we crop 5\% from all sides of all images before computing the scores in testing. 

\subsubsection{3D Detection}
KITTI 3D object detection benchmark comprises 7481 training images and 7518 test images, along with the corresponding point clouds captured around a midsize city from rural areas and highways. KITTI provides 3D bounding box annotations for 3 classes, Car, Cyclist and Pedestrian. Commonly, the training set is divided into training split with 3712 samples and validation split with 3769 samples following that in \cite{ding2020learning}, which we denote as KITTI train and KITTI val, respectively. All models in ablation studies are trained on the KITTI train and evaluated on KITTI val. For the submission of our methods, the models is trained on the 7481 training samples. Each object sample is assigned to a difficulty level, Easy, Moderate or Hard according to the object is bounding box height, occlusion level and truncation.

\subsection{Evaluation Metrics}
\subsubsection{Novel View Synthesis}
To measure the quality of the generated images, we compute the Structural Similarity Index (SSIM), PSNR, and the recently proposed LPIPS perceptual similarity. We use an ImageNet-trained VGG16 model when computing the LPIPS score. 

\subsubsection{Stereo 3D Detection}
We use two evaluation metrics in KITTI-3D, i.e., the IoU of 3D bounding boxes or BEV 2D bounding boxes with average precision (AP) metric, which are denoted as $AP_{3D}$ and $AP_{BEV}$, respectively. Following the monocular 3D detection methods \cite{barabanau2019monocular,ding2020learning,zhang2021objects}, we conduct the ablation study on Car. KITTI-3D uses the  $A P |_{ R40}$ with 40 recall points instead of  $A P |_{ R11}$ with 11 recall points from October 8, 2019. We report all the results in $A P |_{ R40}$.

\subsection{Implementation Details}
\subsubsection{Novel View Synthesis}In the training phase, the number of time steps was set to 1000, and we used an NVIDIA Tesla V100 GPU with 32G of memory, and the batch size was set to 16 with the same pre-trained VQGAN model as the Latent Diffusion model, and 45 epochs were performed in 3 days. For optimization, we use AdamW \cite{adamw} optimizer with $\beta $ (0.9, 0.999), weight decay 0.1 and dropout rate 0.1, and an exponential moving average (EMA) optimizer with a coefficient of 0.9999. In the inference phase, we used 1000 sampling steps for the methods without acceleration and for the methods with acceleration, the sampling steps were method dependent, as described in the ablation experiments.

\subsubsection{Stereo 3D Detection}We use LIGA-Stereo, stereoyolo and stereocenternet as baselines for stereoscopic 3D detection according to the method. we use 2 NVIDIA RTX3090 GPU to train this networks. the LIGA-Stereo batch size is set to 2, the stereoyolo batch size is set to 2 and the stereocenternet batch size is set to 2. We use one model to detect different classes of objects (Car, Cyclist and Pedestrian) simultaneously, and other hyperparameter settings are the same as LIGA-Stereo, YOLOStereo3D and Stereo-CenterNet.

\begin{table*}[htbp]
	\centering
	\scriptsize 
	\caption{Car Localization and Detection. $AP_{BEV}/AP_{3D}$ on \textit{validation} set.}
	\begin{tabular}{c|c|ccc|ccc}
		\toprule
		\multirow{2}[2]{*}{Methods} & \multirow{2}[2]{*}{Reference} &       & $AP_{3D}$  &       &       & $AP_{BEV}$  &  \\
		&       & Easy  &  Moderate & Hard  & Easy  & Moderate & Hard \\
		\midrule
		MonoDIS \cite{simonelli2019disentangling} & ICCV 2019 & 10.37 & 7.94  & 6.40  & 17.23 & 13.19 & 11.12 \\
		AM3D \cite{ma2019accurate} & ICCV 2019 & 16.50 & 10.74 & 9.52  & 25.03 & 17.32 & 14.91 \\
		M3D-RPN \cite{brazil2019m3d} & ICCV 2019 & 14.76 & 9.71  & 7.42  & 21.02 & 13.67 & 10.23 \\
		D4LCN \cite{ding2020learning} & CVPR 2020 & 16.65 & 11.72 & 9.51  & 22.51 & 16.02 & 12.55 \\
		MonoPair \cite{chen2020monopair} & CVPR 2020 & 13.04 & 9.99  & 8.65  & 19.28 & 14.83 & 12.89 \\
		MonoFlex \cite{zhang2021objects} & CVPR 2021 & 19.94 & 13.89 & 12.07 & 28.23 & 19.75 & 16.89 \\
		MonoEF \cite{zhou2021monocular} & CVPR 2021 & 21.29 & 13.87 & 11.71 & 29.03 & 19.70 & 17.26 \\
		GrooMeD-NMS \cite{kumar2021groomed} & CVPR 2021 & 18.10 & 12.32 & 9.65  & 26.19 & 18.27 & 14.05 \\
		CaDDN \cite{reading2021categorical} & CVPR 2021 & 19.17 & 13.41 & 11.46 & 27.94 & 18.91 & 17.19 \\
		DDMP-3D \cite{wang2021depth} & CVPR 2021 & 19.71 & 12.78 & 9.80  & 28.08 & 17.89 & 13.44 \\
		MonoRUn \cite{chen2021monorun} & CVPR 2021 & 19.65 & 12.30 & 10.58 & 27.94 & 17.34 & 15.24 \\
		DFR-Net \cite{zou2021devil} & ICCV 2021 & 19.40 & 13.63 & 10.35 & 28.17 & 19.17 & 14.84 \\
		MonoRCNN \cite{shi2021geometry} & ICCV 2021 & 18.36 & 12.65 & 10.03 & 25.48 & 18.11 & 14.10 \\
		DD3D \cite{park2021pseudo} & ICCV 2021 & 23.22 & 16.34 & 14.20 & 30.98 & 22.56 & 20.03 \\
		PS-im \cite{chen2022pseudo} & CVPR 2022 & 23.74 & 13.81 & 12.31 & 28.37 & 20.01 & 17.39 \\
		\midrule
		Ours-BBDM &       & 20.37 & 13.93 & 13.54 & 28.34 & 20.61 & 22.51 \\
		Ours-View &       & 22.25 & 14.62 & 15.26 & 31.16 & 22.24 & 23.18 \\
		\bottomrule
	\end{tabular}%
	\label{tb}%
\end{table*}%

\begin{table}[htbp]
	\centering
	\scriptsize
	\caption{Performance for Car on KITTI val set at IOU threshold 0.7. The best results are bold, the second best \underline{underlined}.}
	\begin{tabular}{c|ccc}
		\toprule
		\multirow{2}[2]{*}{Methods} &       & $AP_{3D}$  &  \\
		& Easy  &   Moderate & Hard \\
		\midrule
		D4LCN & 22.32 & 16.20 & 12.30 \\
		DDMP-3D & 28.12 & 20.39 & 16.34 \\
		CaDDN  & 23.57 & 16.31 & 13.84 \\
		MonoFlex  & 23.64 & 17.51 & 14.83 \\
		GUPNet & 22.76 & 16.46 & 13.72 \\
		PS-im & 31.81 & 22.36 & 19.33 \\
		PS-fld & \textbf{35.18} & \textbf{24.15} & 20.35 \\
		\midrule
		Ours-BBDM & 30.6  & 23.55 & 20.07 \\
		Ours-View & \underline{32.55}  & \underline{24.06} & \textbf{22.14} \\
		\bottomrule
	\end{tabular}%

	\label{td}%
\end{table}%

\begin{table*}[htbp]
	\centering
	\scriptsize 
	\caption{ Performance for Pedestrian and Cyclist on KITTI test at IOU threshold 0.5. \newline
	The best results are bold, the second best \underline{underlined}.}
	\begin{tabular}{c|ccc|ccc}
		\toprule
		\multirow{2}[2]{*}{Methods} & \multicolumn{3}{c|}{Pedestrian $AP_{3D}/AP_{BEV}$ } & \multicolumn{3}{c}{Cyclist $AP_{3D}/AP_{BEV}$} \\
		& Easy  & Moderate & Hard  & Easy  & Moderate & Hard \\
		\midrule
		D4LCN & 4.55 / 5.06 & 3.42 / 3.86 & 2.83 / 3.59 & 2.45 / 2.72 & 1.67 / 1.82 & 1.36 / 1.79 \\
		MonoPSR  & 6.12 / 7.24 & 4.00 / 4.56 & 3.30 / 4.11 & 8.37 / 9.87 & 4.74 / 5.78 & 3.68 / 4.57 \\
		CaDDN  & 12.87 / 14.72 & 8.14 / 9.41 & 6.76 / 8.17 & 7.00 /9.67 & 3.41 / 5.38 & 3.30 / 4.75 \\
		MonoFlex & 9.43 / 10.36 & 6.31 / 7.36 & 5.26 / 6.29 & 4.17 / 4.41 & 2.35 / 2.67 & 2.04 / 2.50 \\
		GUPNet & 14.95 / 15.62 & 9.76 / 10.37 & 8.41 / 8.79 & 5.58 / 6.94 & 3.21 / 3.85 & 2.66 / 3.64 \\
		PS-im & 8.26 / 9.94 & 5.24 / 6.53 & 4.51 / 5.72 & 4.72 / 5.76 & 2.58 / 3.32 & 2.37 / 2.85 \\
		PS-fld & \textbf{16.95 / 19.03} & 10.82 / 12.23 & 9.26 / 10.53 & 11.22 / 12.80 & 6.18 / 7.29 & 5.21 / 6.05 \\
		PS-fcd & 14.33 / 17.08 & 9.18 / 11.04 & 7.86 / 9.59 & 9.80 / 11.92 & 5.43 / 6.65 & 4.91 / 5.86 \\
		\midrule
		Ours-BBDM & \underline{15.16 / 17.46} & \textbf{12.74 / 14.18} & \textbf{10.83 / 12.77} & \textbf{11.99/ 12.98} & \textbf{8.24 / 8.49} & \textbf{7.85 /8.27} \\
		\bottomrule
	\end{tabular}%

	\label{tb}%
\end{table*}%

\subsection{Single-image-based View Synthesis Results}
\subsubsection{Quantitative Results}To prove the effectiveness of our approach, we conduct a large number of comparative experiments. The compared algorithms include DAM-CNN \cite{xie2016deep3d}, Tulsiani et. al. \cite{tulsiani2018layer}, MPI \cite{tucker2020single} and MINE \cite{li2021mine}. The quantitative experimental results are shown in Table 1. The test resolution of the images of all our approaches is set to $256\times 768$ to make a fair comparison. Our approach is
significantly better than DAM-CNN,Tulsiani et. al., MPI. The PSNR of our approach can surpass SOTA after adopting EMSA and feature-level parallaxaware loss, and the SSIM and LPIPS scores are slightly inferior to SOTA \cite{li2021mine}.

\subsubsection{Qualitative Results }We also qualitatively demonstrate our superior view synthesis performance in Fig. 7. Obviously, Our approach has achieved competitive performance to the state-of-the-art method and synthesizes more realistic images with fewer distortions and artifacts compared with other methods. Compared to \cite{tucker2020single}, we generate more realistic images with lesser artefacts and shape distortions. The visualization verifies our ability to model the geometry and texture of complex scenes.

\begin{figure*}[t]
	\centering
	\includegraphics[width=15.12cm,height=4cm]{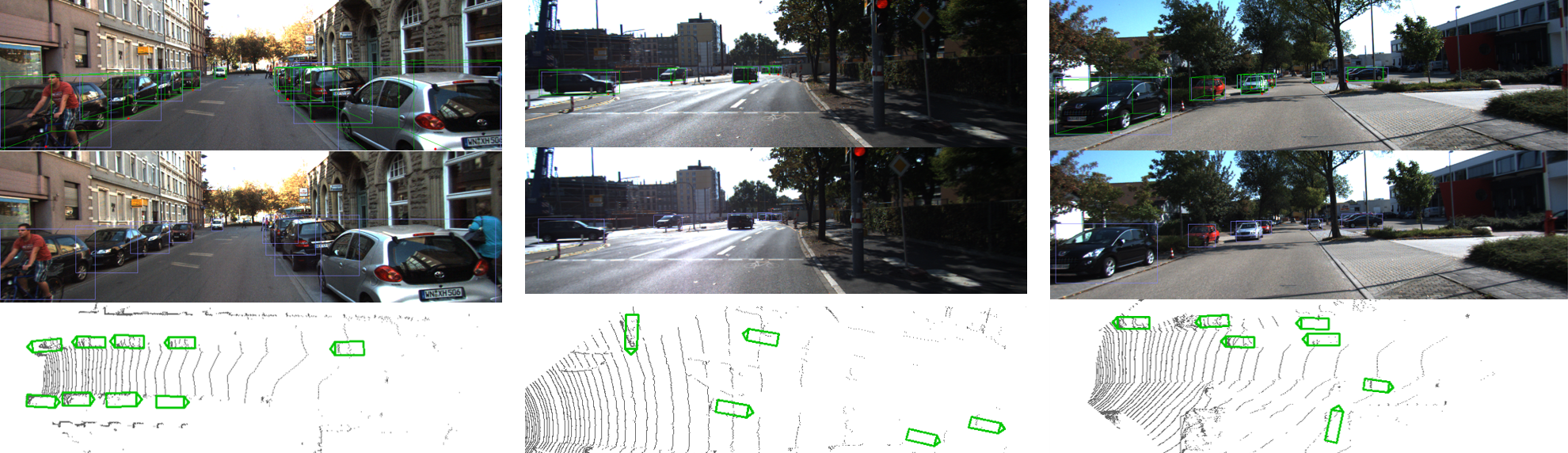}	
	\caption{Quantitative results of multiple scenarios in the KITTI dataset. The first row presents the predicted 3D bounding boxes drawn from the detection results of the left image, the second row depicts the 2D bounding boxes in the right-eye image, and the third row presents the aerial view image.}
	\label{i}
\end{figure*}

\subsection{3D Object Detection Results}
In this section, we evaluate the proposed three pseudo-stereoscopic variants:BBDM, View-operator and one-step generation, on the KITTI test and val sets, and other monocular 3D detectors are compared.

\subsubsection{Quantitative Results}The results reported in Table 2 and Table 3 show that the large interval performance of the method proposed in 3D target detection and 3D positioning is better than all other methods. Even if we only use BBDM as the basic diffusion model, the performance of the two tasks with 0.7 with the IOU threshold can be significantly better than the most advanced method, such as Monorcnn and DD3D. Generally speaking, better image generation can improve the performance of 3D target detection and positioning. We can see that the advantages of the View diffusion model are more significant compared to BBDM. Due to the same super reassembly, such as learning rate, average pixel, backbone network, and the size of the priority box, the View method has better performance, indicating that the View structure has better generalization capabilities for 3D target detection.

When the IOU threshold is 0.7, compared with our baseline method PS-IM, it is slightly lower in simple samples, but the performance of 3D target detection and positioning tasks in suffering and medium samples has greatly improved, about 1 in 1, about 1 -2, these improvements prove the effectiveness of the method. We attribute small gaps on simple samples to limited constraints. Remember, our method directly uses the diffusion model to generate the right figure. Although we have added image translation as a constraint, compared with the depth diagram and geometric priority, the formation method is not completely controllable. Without matching the texture, the background and the obscure object inevitably bring interference to the new perspective generation. The Convnext-UNET proposed in this article can alleviate this problem, which has been proven in ablation research, but it is not perfect.

In addition, we reported the evaluation results of the Kitti verification set. As shown in Table III, the method is obviously better than our previous methods D4LCN and the latest methods, such as DDMP-3D, Caddn, Monoflex, and Gupnet. Compared with the baseline method PS-IM and PS-FLD, there is only a weak gap in simple and medium, and two points are improved in difficulties.

\subsubsection{Qualitative Results }We present the qualitative results of a number of scenarios in the KITTI dataset in the Figure 8. We present the corresponding stereo box, 3D box, and aerial view on the left and right images. It can be observed that in general street scenes, the proposed SC can accurately detect vehicles in the scene, and the detected 3D frame can be optimally aligned with the LiDAR point cloud. It also detected a few small objects that were occluded and far away.
\subsection{Ablation Study}
In this part, we will present the ablation study to verify the effectiveness of some important components of the proposed method. To investigate the effects of different components of our approach, we set up several different versions, as shown below:

\begin{itemize}
	\item[$\bullet$]Pedestrian and Cyclist 3D detection results.
	\item[$\bullet$]Whether to speed up sampling.
	\item[$\bullet$]Setting of the hyperparameter s in BBDM.
	\item[$\bullet$]Latent+U-NET.
	\item[$\bullet$]Latent+ConvNeXt-UNet.
	\item[$\bullet$]Image size.
	\item[$\bullet$]Different stereo detectors.
	\item[$\bullet$]Different optimizers.
	\item[$\bullet$]Performance of SSIM Loss.

\end{itemize}


\noindent{\bf Pedestrian and Cyclist 3D detection results.} In the KITTI object detection benchmark, the training samples of \textit{Pedestrian} and \textit{Cyclist} are limited; hence, it is more difficult than detecting \textit{car} category. Because most image-based methods do not exhibit the evaluation results of \textit{Pedestrian} and \textit{Cyclist}, we solely report the available results of the original paper. We present the pedestrian and cyclist detection results on KITTI \textit{validation} set in Table 4, SVDM achieves the best detection results except for pedestrian simple samples.

The remaining ablation experiments were temporarily not completed due to time reasons.

\section{CONCLUSION AND FUTURE SCOPE}
We propose SVDM, a new pseudo-stereo image 3D object detection method, and we solve the new single-view view synthesis problem as an image-to-image translation problem by combining it with the latest diffusion model. The proposed SVDM achieves the best performance without geometric priors, depth estimation and LIDAR monitoring, demonstrating that image-based methods have great potential in 3D.

However, the proposed framework does not allow end-to-end training. Therefore, we can try to further refine and simplify the framework by end-to-end training while guaranteeing the detection performance. Another major limitation of the method is that the new view generation falls short of the SOTA method, and in the future, we will further add new components to this method to further improve the accuracy of the new view generation task.

\section*{Acknowledgments}
This research work is supported by the Big Data Computing Center of Southeast University.
 
\bibliography{tip}{}
\bibliographystyle{IEEEtran}

\vfill

\end{document}